\documentclass[10pt,journal,compsoc]{IEEEtran}
\ifCLASSOPTIONcompsoc
  \usepackage[nocompress]{cite}
\else
  \usepackage{cite}
\fi
\ifCLASSINFOpdf
\else
\fi
\usepackage{graphicx}
\usepackage{algorithm}
\usepackage{bm}
\usepackage{color}
\usepackage{multirow}
\usepackage{ifthen}
\usepackage{bbm}
\usepackage{wrapfig}
\usepackage{amsmath}
\usepackage{amssymb}
\graphicspath{{./figures/}}

\renewcommand{\paragraph}[1]{\vspace{0.1cm}\noindent\textbf{#1}\quad}

\newcommand{\method}[1]{\ifthenelse{\equal{#1}{full}}{self-SUPervised REMEdy}{SUPREME}}

\newcommand{\best}[1]{{\textbf{\color{red}#1}}}
\newcommand{\second}[1]{{\underline{\color{blue}#1}}}

\begin{document}
%
\title{Unsupervised Transfer Learning with Self-Supervised Remedy}
%
%
%
%

\author{Jiabo~Huang, Shaogang~Gong
\thanks{Jiabo Huang and Shaogang Gong are with the School of Electronic Engineering and Computer Science, Queen Mary University of London, UK. E-mail: \{jiabo.huang, s.gong\}@qmul.ac.uk.}}

\IEEEtitleabstractindextext{%
\begin{abstract}
Generalising deep networks to novel domains 
without manual labels
is challenging to deep learning.
This problem is intrinsically difficult due to unpredictable changing nature of
imagery data distributions in novel domains.
Pre-learned knowledge does not transfer well 
without making strong assumptions about 
the learned and the novel domains.
Different methods have been
studied to address the underlying problem based on different assumptions, 
\textit{e.g.}
from domain adaptation to zero-shot and few-shot learning.
In this work,
we address this problem by 
transfer clustering
that aims to learn a {\em discriminative} latent space
of the {\em unlabelled} target data in a novel domain by
knowledge transfer from labelled related domains.
Specifically,
we want to leverage relative (pairwise) imagery information,
which is freely available and intrinsic to a target domain,
to model the target domain image distribution characteristics
as well as
the prior-knowledge learned from related labelled domains
to enable more discriminative clustering of unlabelled target
data. 
Our method mitigates nontransferrable prior-knowledge by self-supervision,
benefiting from both transfer and self-supervised learning.
Extensive experiments on four datasets for image clustering tasks
reveal the superiority of our model over the state-of-the-art
transfer clustering techniques.
We further demonstrate its competitive transferability
on four zero-shot learning benchmarks.
\end{abstract}

}

\maketitle

\IEEEdisplaynontitleabstractindextext

%
\IEEEpeerreviewmaketitle

\IEEEraisesectionheading{\section{Introduction}\label{sec:intro}}

\IEEEPARstart{D}{espite} the remarkable progress advanced by deep convolutional neural
networks (CNNs) on computer vision in recent years~\cite{goodfellow2016deep,lecun2015deep},
the \textit{i.i.d} 
(independent and identically distributed) 
assumption 
widely held by most of the deep learning models
restricts their transferability and usability in novel domains without
additional labelled training data.
In general, realistic data of interest usually have different and
unknown distributions in novel domains (domain shift).
It is both labour-intensive to manually annotate sufficient data
and computationally expensive to retrain a model in every new target domain.
%
To overcome this fundamental problem, \textit{transfer learning}
(TL)~\cite{torrey2010transfer} has been widely studied,  
which aims to 
leverage the knowledge gained from one domain
to help acquire a better understanding of 
other related domains.
%
Moreover, only unlabelled data are mostly available at scale,
therefore, 
\textit{unsupervised} transfer learning~\cite{pan2009transfer} 
serves as a more natural solution in the deep learning context
and has attained increasing attention~\cite{lee2019DTA,han2019DTC,ye2019learning}.

\begin{figure*}[ht]
\centering
\includegraphics[width=0.8\linewidth]{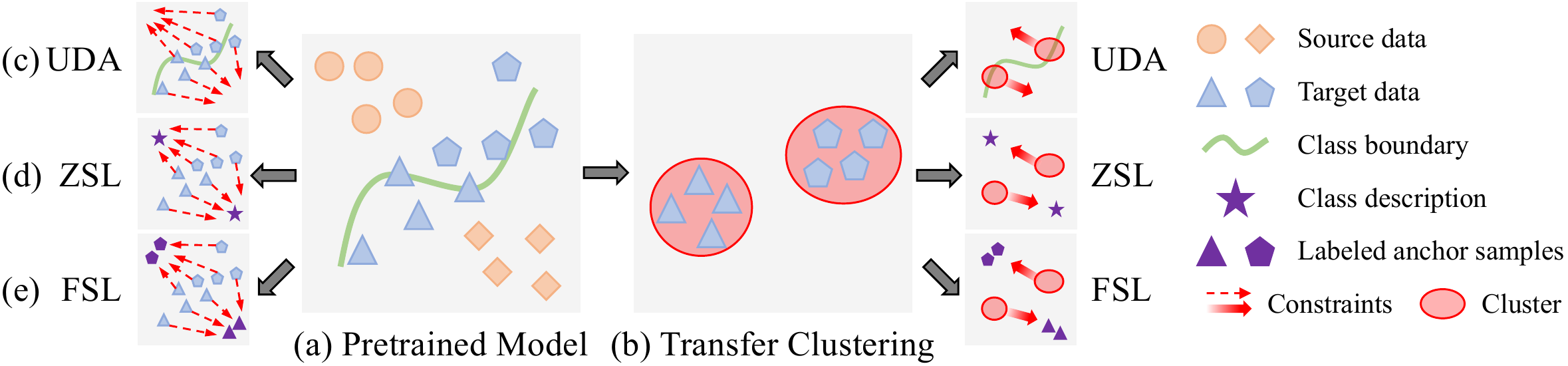}
\caption{
Popular solutions to unsupervised transfer learning.
\textbf{(a)} Model pretrained on source domain with human annotations.
\textbf{(b)} Cluster analysis on target domain using the prior from source domain.
\textbf{(c)} Unsupervised domain adaptation (UDA) assumes that 
source and target data are drawn from the same label space.
\textbf{(d)} Zero-shot learning (ZSL) represents
both the seen and unseen classes by 
exploiting pretrained word embedding in a text space.
\textbf{(e)} A few labelled anchor samples are available 
as the prototype of novel classes in few-shot learning
(FSL).
}
\label{fig:transfer_learning}
\end{figure*}
%
Recent efforts on unsupervised transfer learning 
mainly start from three different assumptions.
To deal with target data sampled from arbitrary distributions,
Unsupervised Domain Adaptation (UDA)~\cite{ben2010da} assumes that
they are drawn from the same label space as the source data
(Fig~\ref{fig:transfer_learning} (c)).
In this case,
the trained model can be adapted to the target domain by
refining decision boundaries that are optimal for both distributions.
However, the largest imagery database~\cite{russakovsky2015imagenet} available today
contains samples from only 1,000 classes,
which is far less than the number of object categories known by human
(\textit{e.g.} in language), 
let alone the countless unknown ones. 
Therefore, assuming all the newly collected data are from the known classes 
is sometimes impractical.
To relax such assumptions,
Zero-Shot Learning (ZSL)~\cite{lampert2009zsl} 
and Few-Shot Learning%
\footnote{
Due to the efforts made by FSL 
on cross-domain knowledge transfer
with insufficient labelled data supervision,
we discuss it in the wider context of unsupervised transfer learning
although it is not strictly unsupervised.
}
(FSL)~\cite{miller2000fsl} 
are introduced, 
in which the target data are from novel classes
that are
either {\em unseen} or {\em weakly-seen} during model training.
Importantly, additional information on the relationships between seen
and unseen or weakly-seen classes are required. That is, in ZSL novel
classes (unseen) must be {\em known semantically} in a word vector space in a sense that they
are in proximity to the seen classes semantically, \textit{i.e.} interpretable in a text
space (Fig~\ref{fig:transfer_learning}~(d)). In FSL, novel
classes are weakly-seen by having a few labelled samples 
(usually 1$\sim$5 samples per class) as the anchors
(Fig~\ref{fig:transfer_learning}~(e)). Moreover, novel classes are also
implicitly assumed to be \textit{known} and in proximity to the seen classes in a visual
feature space, \textit{i.e.} interpretable visually.
As it is hopeless to define all the categories 
even in relatively restricted settings~\cite{han2019DTC},
it remains an open question as how to generalise a trained CNN model
to the data from {\em real} novel classes, that is, both unseen and \textit{unknown}.

Whilst existing unsupervised transfer learning methods make different
assumptions that limit their usability to different scenarios,
they do share the same objective of mapping images with
no/limited labelled training data (target data) 
into a discriminative latent space,
in which samples from the same classes are closer than those from
different classes (Fig~\ref{fig:transfer_learning}~(b)).
We observe this objective being identical to that of cluster analysis,
and we consider that the key challenge of 
unsupervised transfer learning is to
leverage knowledge from human supervision on related source data
for more discriminative clustering of unlabelled target data, 
\textit{i.e.} Transfer Clustering (TC)~\cite{han2019DTC}.
Transfer Clustering is fundamentally more challenging than
the abovementioned settings 
due to the fewer assumptions it makes on target domains.
We empirically show that
our proposed TC method
can attain competitive knowledge transferring ability
on ZSL tasks, whilst 
it
provides a more generic solution to
unsupervised transfer learning than contemporary ZSL models.

In this work,
we introduce a novel transfer clustering method called
\textit{\method{full}} ({\method{abbr}). 
%
The motivation of the \method{abbr} model is that,
the supervision constructed according to
the prior-knowledge acquired from related 
domains,
called \textit{transferred supervision},
is not always sufficient for all the target samples
due to distribution shift and class non-proximity (discrepancy).
Therefore,
additional complementary supervision is necessary 
for learning a model to better describe the target distribution.
%
Specifically,
by exploring the auxiliary learning principle~\cite{caruana1997multitask},
\method{abbr} jointly learns by transferred supervision and {\em self-supervision},
which are individually adjusted on each (unlabelled) target image sample
according to
the estimated confidence of the prior-knowledge on it.
The transferred supervision is only considered reliable 
for the samples that falling within the overlapping areas of both
source and target distributions
whilst self-supervision is introduced to enhance
the weak or ambiguous transferred supervision
for the remaining samples.
Both types of supervision are formulated in a single objective
function harmoniously to enable end-to-end model learning.
%

The contributions of this work are three-folded:
\textbf{(1)}
We propose the idea of
exploiting self-supervision as the remedy for unreliable unsupervised transfer learning
so to yield more discriminative modelling of the target distributions.
To our best knowledge,
this is the first attempt at 
jointly using related domain prior-knowledge and novel target domain self-supervision
for clustering by deep neural networks.
\textbf{(2)}
We formulate a \textit{\method{full}} (\method{abbr}) method
for transfer clustering that enables an effective implementation of
leveraging related source domain prior-knowledge
whilst simultaneously mitigating the negative impact of ambiguous transferred supervision
by self-supervised learning in a target domain.
\textbf{(3)}
We empirically show that transfer clustering by \method{abbr} is a
more generic solution to unsupervised transfer learning than ZSL.

The superiority of the proposed method
over a wide range of existing unsupervised transfer learning techniques
is shown by extensive experiments 
on both transfer clustering and ZSL tasks
using 8 benchmarks:
CIFAR10/100~\cite{krizhevsky2009cifar}, 
SVHN~\cite{netzer2011svhn},
ImageNet~\cite{russakovsky2015imagenet},
CUB~\cite{WelinderEtal2010cub},
SUN~\cite{patterson2012sun},
FLO~\cite{nilsback2008flo}
and AWA2~\cite{xian2018awa}.

\section{Related Work}\label{sec:literature}

Instead of focusing narrowly on transfer clustering,
we consider the bigger picture of unsupervised transfer learning.
This is partly because our method of exploring self-supervised learning
is closely related to unsupervised transfer learning by training deep networks
with insufficient labelled data supervision. 
%
%

\paragraph{\bf Unsupervised Transfer Learning.}
Unsupervised Domain Adaptation (UDA)
has been widely studied
to transfer knowledge across different data distributions.
Critically, by assuming all the domains 
are sharing the same label space,
UDA learns to either align feature distributions%
~\cite{ganin2014unsupervised,tzeng2017adversarial}
or map the relationships between 
the decision boundaries and feature representations so that
the boundaries are valid for both domains%
~\cite{lee2019DTA,saito2018maximum}.
However, 
UDA is not always practical 
as it cannot enumerate all the categories for model training
let alone exhaustively collecting and annotating the data.
Alternatively,
Zero-shot Learning (ZSL) and Few-shot Learning (FSL)
have drawn increasing attention,
in which target data of novel classes are unseen or insufficiently-seen during model training.
It is intrinsically challenging to generalise CNN models 
to unseen categories.
To overcome this problem,
both learning tasks hold an assumption that
the novel classes are \textit{known} according to their pre-defined relationships with the
seen classes.
In ZSL~\cite{song2018qfsl,chao2016empirical,lampert2009zsl},
all the seen (source) and unseen (target) classes 
are described by a common representational space, 
\textit{e.g.} attribute space,
so that data from unseen classes can be classified
by novel combinations of attributes which are building
blocks shared with the seen classes. 
Different from ZSL,
FSL~\cite{ye2019learning,miller2000fsl} assumes that 
a few anchor samples with labels are available for each novel class as
its prototype (so strictly not unseen), and the novel classes and seen
classes are in proximity in their distributions.
In this case, 
the target data can be classified 
according to their distances to different anchor samples with the
assistance of the distributions of seen classes.
Whilst standard ZSL and FSL assume that the target label space is completely disjoint to
the seen label space {\em in test}, Generalised ZSL/FSL
(GZSL/GFSL)~\cite{chao2016empirical,ye2019learning} consider that
target samples possibly consist of both seen and novel (unseen) classes.
Transductive ZSL/FSL (tZSL/tFSL)~\cite{song2018qfsl,nichol2018tfsl} 
assume further that unlabelled target samples are also available for model training.
In contrast to ZSL/FSL, transfer clustering is 
a more challenging problem as it makes fewer assumptions on the target
label space in relation to the seen classes.
%

\paragraph{\bf Transfer clustering.} Han \textit{et al.}~\cite{han2019DTC} first introduced
the task of transfer clustering formally,
which aims to 
jointly learn {\em both} the representations and 
the decision boundaries of unlabelled target data 
with the help of the labelled data from related domains.
They construct the initial clustering solution
by applying K-means upon the feature representation produced by a pretrained model
and learn to sharpen the initial assignment distribution of each sample.
There are other attempts at transferring knowledge 
across domains for achieving learning tasks that are similar to transfer clustering.
KCL~\cite{hsu2017KCL} and MCL~\cite{hsu2019MCL} 
are based on a Constrained Clustering Network (CCN),
which learns to transfer pairwise similarities
from a source to a target domain
so that the cross-domain and cross-task transfer learning
are decoupled. This aims to reduce the model learning complexity.
Another method called Centroid Networks~\cite{huang2019centroid} is proposed to
jointly learn data embeddings and clustering
using the Sinkhorn K-means algorithm.
Due to distribution shift and discrepancy in label space proximity,
the clustering of some target samples during model training will not yield
sufficient supervision from the prior-knowledge of the source domain,
therefore, poorly modelled.
We aim to solve this problem by self-supervision
so that the model can better explain the target distribution.

\paragraph{\bf Self-supervised Learning.}
Self-supervised learning is a general concept that explores inherent information in visual data
without labelling in order to construct pseudo-supervision in model training.
The formulation of self-supervision has many manifestations.
One popular approach is to explore pixel information that trains a
model to reconstruct the data distributions, known as a generative
model, \textit{e.g.} the Restricted Bolzmann Machines (RBMs)~\cite{tang2012RBM,lee2009dbn},
Autoencoder~\cite{vincent2010sae,ng2011autoencoder},
and Generative Adversarial Networks (GANs)~\cite{radford2016dcgan,donahue2016bigan}.
Data perturbation is another approach to construct supervision by
introducing positive sample pairs%
~\cite{dosovitskiy2015discriminative,wu2018instance,huang2018and,huang2020pad}.
Self-supervision can also be formulated according to
spatial context~\cite{gidaris2018rotnet,doersch2015context,noroozi2016jigsaw}, 
and colour distributions~\cite{zhang2016color,zhang2017splitbrain}.
Although many self-supervised learning techniques show
the potential to learn from unlabelled data,
it is clearly more beneficial to also explore any prior-knowledge 
when available.
To that end, we wish to augment transfer learning by self-supervised learning.

\section{Transfer Clustering}\label{sec:method}

Given $N^s$ image-label pairs $\{I^s, y^s\}_{i=1}^{N^s}$ 
drawn from a source label space $\mathcal{S} = \{1, 2, \cdots, K^s\}$
and $N^t$ target images $\{I^t\}_{i=1}^{N^t}$ 
from $\mathcal{T} = \{1, 2, \cdots, K^t\}$ 
where $\mathcal{S} \neq \mathcal{T}$.
In transfer clustering,
no class label is annotated on target images.
The objective is to
jointly learn a feature representation of target samples 
$f_\theta: I^t \rightarrow x^v$
and the probabilities that 
they belong to each of $K^t$ clusters
$f_\delta: x^v \rightarrow p$
so that the target samples of the same class labels 
are more likely to be allocated into the same partitions.
%
Due to the absence of labelling in the target domain,
the target distribution is agnostic, 
so as the discrepancy between it and the source distribution.
It is therefore necessary 
to consider not only an effective way for source domain knowledge transfer
in target sample clustering, but also how to identify and deal with those target samples having
insufficient (weak or ambiguous) support from the source domain prior-knowledge.
This is intrinsically challenging
as the arbitrarily complex appearance patterns 
and variations exhibited in the imagery data
usually lead to 
intricate relationships between source domain and
target domain distributions. 
%
\begin{figure}[t]
\centering
\includegraphics[width=1.0\linewidth]{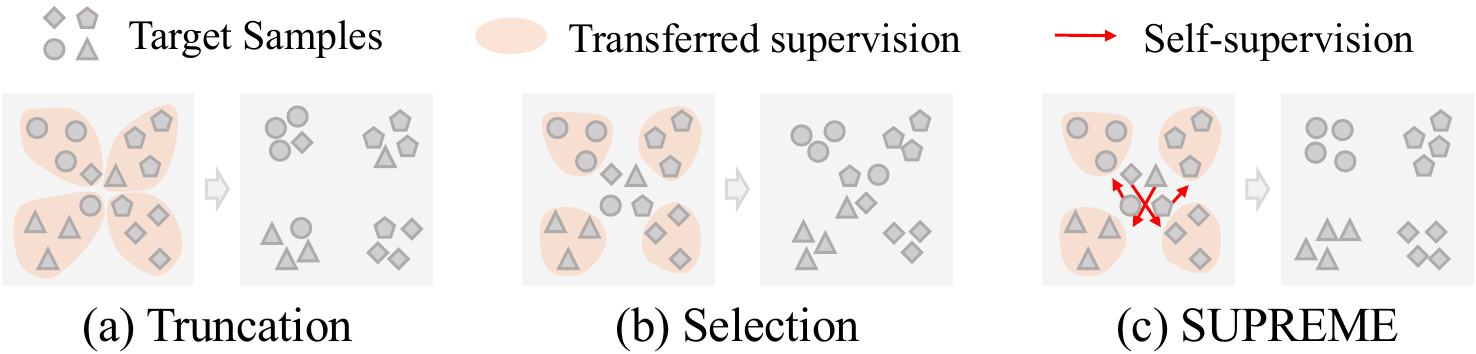}
\caption{
Illustration of different strategies 
to deal with ambiguous transferred supervision.
\textbf{(a)}
Converting any ambiguous transferred supervision into the determined supervision 
by nearest neighbour assignment.
\textbf{(b)} 
Only utilising target samples having confident transferred
supervision, neglect the rest. 
\textbf{(c)}
Augmemt target samples of insufficient transferred supervision 
by self-supervision.
}
\label{fig:ambiguous_supervision}
\end{figure}

To address this problem,
we formulate a \textit{\method{full}} (\method{abbr}) method.
It shares the spirit of auxiliary learning
that trains a model jointly by transferred knowledge and self-supervision.
Our key idea is that,
it can be error-prone and rather arbitrary to truncate the ambiguous supervision 
and convert them mandatorily to be determined
(Fig~\ref{fig:ambiguous_supervision} (a)),
\textit{e.g.} converting the assignment probabilities (soft-labels)
to pseudo (hard) labels according to the nearest cluster.
We consider it is more consistent to
replace those supervisions by 
the intrinsic information encoded in data.
\textit{i.e. self-supervison}
(Fig~\ref{fig:ambiguous_supervision}~(c)).
Although self-supervision does not rely on prior-knowledge unlike the
truncated transferred supervision~\cite{hsu2017KCL,hsu2019MCL}, 
it can minimise the misleading effect of 
applying nontransferrable knowledge to the target domain.
Moreover, \method{abbr} method differs significantly from
the selection strategy~\cite{xie2016dec,han2019DTC}.
The latter neglects samples 
having ambiguous transferred supervision, therefore, introducing bias in model training
that can lead to less discriminative feature space
(Fig~\ref{fig:ambiguous_supervision}~(b)).
The effectiveness of the proposed \method{abbr} model
on a wide range of benchmarks demonstrates its ability to utilise
target samples with insufficient prior-knowledge supervision, rather
than ignore them.
An overview of \method{abbr}
is depicted in Fig~\ref{fig:pipeline}. 
\begin{figure*}[ht]
\centering
\includegraphics[width=0.90\linewidth]{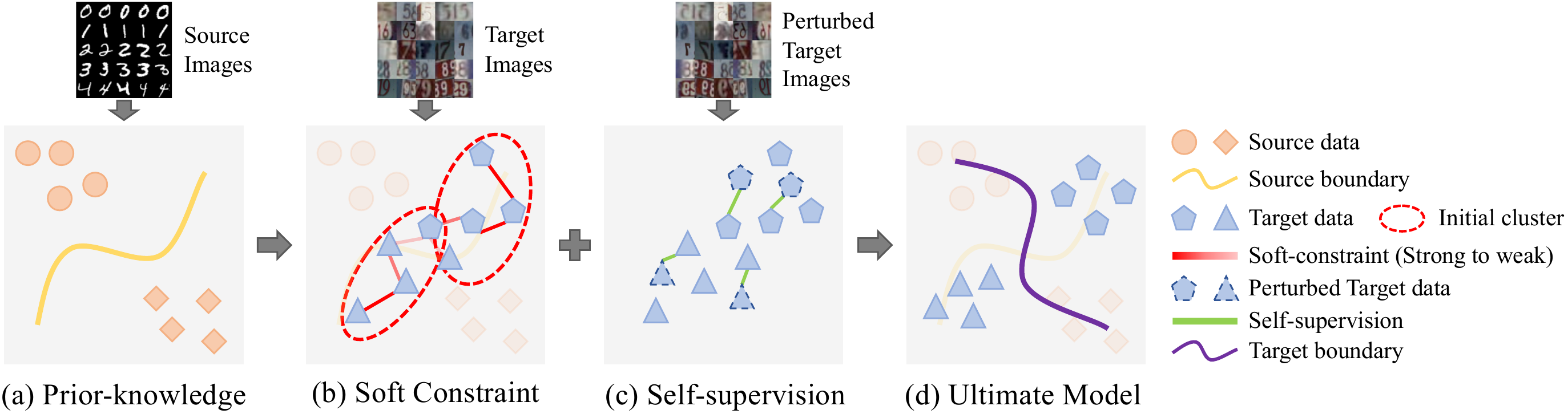}
\caption{
Overview of the proposed \textit{\protect\method{full}} (\protect\method{abbr}) method 
for transfer clustering.
\textbf{(a)} 
We first acquire the prior-knowledge from source domain
by training with manual labels.
\textbf{(b)}
K-means is then applied upon the resulting feature representations
to construct the soft constraints between each pair of target samples,
whose confidence is measured by a novel joint-entropy based metric.
\textbf{(c)}
Self-supervision is then introduced to make up for
the ambiguous constraints.
\textbf{(d)} By jointly trained with the prior-knowledge and self-supervision,
our model finally learns 
a discriminative latent space
as well as the decision boundaries for the target data.
}
\label{fig:pipeline}
\end{figure*}

\subsection{Self-supervised Knowledge Transfer}
We start from how to
construct the initial clustering solution
with the help of a model pretrained on a source domain
to transfer prior-knowledge.
Given a model $\tilde{f}_\theta$ 
which is pretrained by supervised learning on source data $\{I^s, y^s\}_{i=1}^{N^s}$,
transfer learning assumes that 
there is some transferrable common knowledge 
shared by the source and target domains.
As such,
we yield an initial representation of target images $I^t$
by feeding them into $\tilde{f}_\theta: \tilde{x}^t = \tilde{f}_\theta(I^t)$.
An initial clustering of $\tilde{x}^t$ is computed by any standard
technique (\textit{e.g.} K-means). By separating $\tilde{x}^t$ into $K^t$ groups,
we are assuming that all the prior-knowledge of the source domain
are applicable to the target domain by this initial clustering solution.
However, this assumption is not always true due to the distribution shift
and discrepancy in label space proximity.
To address this problem, we formulate the transfer clustering as a
constrained clustering task in which the constraints are formed by
pairwise similarities between target samples determined by their initial assignment and
weighted by 
the estimated confidence.
%

\paragraph{\bf Soft constraints construction.}
Given the representations $\tilde{x}^t$ and 
the $K^t$ clusters centroid $\{c\}_{i=1}^{K^t}$,
we measure the probability $\tilde{p}(c_j|I^t_i)$ that sample $I^t_i$ 
belongs to cluster $c_j$ by computing a student's $t$-distribution 
following~\cite{xie2016dec,han2019DTC}:
\begin{equation}
\tilde{p}(c_j|I^t_i) = \frac{(1 + \lVert \tilde{x}^t_i - c_j \rVert^2/\alpha)^{-(\alpha + 1)/2}}
{\sum_{j'=1}^{K^t}(1 + \lVert \tilde{x}^t_i - c_{j'} \rVert^2/\alpha)^{-(\alpha + 1) / 2}}
\label{eq:assignment}
\end{equation}
Parameter $\alpha$ in Eq~\ref{eq:assignment} is the freedom of student's $t$-distribution 
and is set to 0 in our implementation. 
We denote $\tilde{p}(c_j | I^t_i)$ by $\tilde{p}_{i,j}$ for brevity.
With the initial assignment probabilities,
we then estimate how likely two samples 
are from the same class
by the inner product between their initial assignment distributions:
\begin{equation}
\tilde{r}_{i,j} = \tilde{p}^T_i \cdot \tilde{p}_j = 
\sum_{k=1}^{K^t}\tilde{p}_{i,k}\cdot\tilde{p}_{j,k}
\label{eq:constrains}
\end{equation}
By taking the joint probabilities of sample pairs 
as the measurement of positive relation,
we explore both the \textit{global} and \textit{local}
structures of the pretrained feature space. 
The value of $\tilde{r}_{i,j}$ reaches its maximum $\tilde{r} \rightarrow 1$ only when
two samples 
are both close to each other (local structure)
and close to the same cluster's centroid (global structure).
In which case, they are considered ``confidently positive''.
Otherwise,
the pairwise relation is either 
ambiguous (neither samples is close to any cluster centroid) or 
``confidently negative'' (two samples are close to different cluster
centroids), resulting in the positive probability between them 
becomes the minimum 
$\tilde{r} \rightarrow 0$.
We then train a CNN model with the following soft constraints
to encourage target pairs with large $\tilde{r}_{i,j}$
to be assigned into the same groups:
\begin{equation}
\mathcal{L}_\text{clu} = -\frac{1}{n^2}\sum_{i=1}^n\sum_{j=1}^n\tilde{r}_{i,j}\log r_{i,j},
\quad
r_{i,j} = \sum_{k=1}^{K^t}p_{i,k} \cdot p_{j,k} \\
\label{eq:cluster_loss}
\end{equation}
where $n$ is the mini-batch size and 
$r_{i,j}$ is the up-to-date positive probability of 
the sample pair consisting of $I^t_i$ and $I^t_j$.
To optimise $\mathcal{L}_\text{clu}$, 
the assignment distribution $p$ is encouraged to be ``sharp'' (one-hot in extreme case).
Hence,
even the samples initially are ambiguous
will gradually shift together with other samples of high visual similarity 
to the most plausible clusters.


In the formulation of $\mathcal{L}_\text{clu}$,
ambiguous and confidently negative sample pairs
hold similar low probabilities
to be assigned into the same clusters.
However,
comparing with confidently negative pairs, 
samples of ambiguous pairs are likely positive.
It means that the prior-knowledge transferred to the confidently negative pairs
are more reliable than that to the ambiguous pairs.
Instead of learning from all the prior in equal importance,
the model should 
be encouraged to focus on
the transferrable parts.
To that end,
we assume that target samples near the initial clusters' centroids
are able to form confident pairwise relations
while those close to the decision boundaries cannot.
We quantify the confidence of the pairwise relations
according to the joint entropy of initial assignment distributions:
\begin{equation}
\begin{gathered}
H(I^t_i, I^t_j) = -\sum_{k=1}^{K^t}\tilde{p}_{i,k}\tilde{p}_{j,k}\log \tilde{p}_{i,k}\tilde{p}_{j,k},\quad
H_\text{max} = \log (K^t)^2 \\
w_{i,j} = \frac{\text{exp}((H_\text{max} - H(I^t_i,I^t_j) / (H_\text{max} \cdot \tau))}{\sum_{i',j'}^n\text{exp}((H_\text{max} - H(I^t_{i'},I^t_{j'})) / (H_\text{max} \cdot \tau)}
\label{eq:certainty}
\end{gathered}
\end{equation}
where $\tau$ is the temperature 
that controls the concentration of the confidence distribution;
$H(I^t_i,I^t_j)$ is the joint entropy of $\tilde{p}_i$ and $\tilde{p}_j$;
$w_{i,j}$ is the normalised confidence of 
constraint $\tilde{r}_{i,j}$.
The overall penalty of a mini-batch will then be determined by 
the weighted sum 
instead of the average in Eq~\ref{eq:cluster_loss}:
\begin{equation}
\mathcal{L}_\text{clu} = -\sum_{i=1}^{n^t}\sum_{j=1}^{n^t}w_{i,j}\tilde{r}_{i,j}\log r_{i,j}
\label{eq:weighted_cluster_loss}
\end{equation}
The confidence strategy measures 
the reliability of prior-knowledge from the source domain
in terms of different target samples,
which mitigates the misleading effects
caused by applying nontransferrable prior to target domain.

\paragraph{\bf Self-supervised Remedy.}
As determined by the cost function $\mathcal{L}_\text{clu}$,
samples falling into an ambiguous area of the initial feature space
make necessarily less contribution to model learning as they are given smaller weights.
However,
those samples actually play a significant role 
to learn a more discriminative feature space.
Due to the absence of ground-truth labels
and the ineffectiveness of source domain prior-knowledge on these ``hard'' samples,
the information we can leverage instead
for additional supervision on model learning 
are intrinsic characteristics of the target images.
Inspired by recent unsupervised learning ideas~\cite{xu2019iic,wu2018instance,dosovitskiy2015discriminative},
we formulate 
the self-supervision 
to augment transfer learning
by being invariant to data perturbations.
By applying random image transformations $g(\cdot)$ on the original data,
the positive probability $r_{i,j}$ in Eq~\ref{eq:weighted_cluster_loss}
is computed according to the assignment distribution of sample $I^t_i$ and 
that of $g(I^t_j)$: $r_{i,j} = \sum_{k=1}^{K^t}p_{i,k}\cdot g(p_{j,k})$
where $g(p_j) = f_\delta(f_\theta(g(x_j^t)))$.
Moreover,
we set $\tilde{r}_{i,i} = w_{i,i} = 1 \forall i \in [1, n]$
because the perturbed copies of images are 
certainly with the same class labels as the originals. 
In this case, 
the two supervisions are integrated harmoniously
by our soft-constrained formulation (Eq~\ref{eq:weighted_cluster_loss}).

\subsection{Model Training}
Beyond the self-supervised transfer objective $\mathcal{L}_\text{clu}$,
our model is also trained with several regularisations 
to refrain from degenerated solutions.
The training objective of cluster analysis encourages 
the maximisation of intra-cluster compactness and inter-cluster diversity,
hence,
the model can possibly collapse by
assigning all the samples 
into one single cluster.
Therefore,
we introduce a \textit{balance regularisation} on cluster size:
\begin{equation}
\mathcal{L}_\text{balance} = \log K^t + \sum_{k=1}^{K^t} s_k \log s_k,\quad 
s_k = \frac{1}{n} \sum_{i=1}^{n} p_{i,k}
\end{equation}
where $s_k$ is the approximated size of the $k$-th cluster
and the maximal entropy $\log K^t$ 
is added to ensure positive regularisation values.
We train the model to minimise the negative entropy of 
the approximated cluster size distribution
so as to avoid extremely imbalanced distributions.
Moreover, 
to avoid learning trivial data representations,
we map the visual features to
a common factor space shared by both the source and target domains
with the motivation that
recognition should be conducted on the space
where each factor can be interpreted as a latent attribute%
~\cite{fu2013learning,rastegari2012attribute,chang2019dlstl}.
Specifically,
given the visual feature $x^v$ of sample $I$ 
from either domains,
we project it to the latent factor space produced by a linear layer $f_\omega$
and activate it by the non-linear \textit{Sigmoid} function 
$x^a=\sigma(f_\omega(x^v))$.
An element-wise \textit{binary regularisation} is then applied on the factor space:
\begin{equation}
\mathcal{L}_\text{attr} = -\frac{1}{n \times D}\sum_{i=1}^n\sum_{j=1}^Dx^a_{i,j}\log x^a_{i,j} + (1 - x^a_{i,j})\log(1 - x^a_{i,j})
\label{eq:common_space}
\end{equation}
where $D$ is the dimension of the attribute representation 
$x^a \in \mathbb{R}^D$.
The binary regularisation attains its minimum $\mathcal{L}_\text{attr} \rightarrow 0$
when $x^a \in \{0, 1\}^D$.
Afterwards,
$x^a$ will be fed into the domain-corresponding classifier
to predict the assignment distribution.
The prediction of target samples 
is supervised by 
our proposed objective function (Eq~\ref{eq:weighted_cluster_loss})
while that of source data is
by the conventional cross-entropy loss with the provided labels on source domain:
\begin{equation}
\mathcal{L}_\text{xent} = -\frac{1}{n}\sum_{i=1}^n\sum_{j=1}^{K^s}\mathbbm{1}[j = y^s_i] \log p_{i,j}
\label{eq:xent}
\end{equation}
The $\mathbbm{1}$ denotes the indicator function 
which equals to $1$ \textit{iff} $j$ is the ground-truth label $y^s_i$,
otherwise $0$. 
Although our ultimate goal is on target domain,
jointly training with the classification task on source domain 
can also be taken as a \textit{multi-tasks regularisation}
to avoid learning trivial representation 
as well as 
the well-known catastrophic forgetting problem 
in transfer learning~\cite{mccloskey1989catastrophic,goodfellow2013empirical}.

\section{Experiments}\label{sec:exp}

\paragraph{\bf Datasets.}
Evaluations of the proposed \method{abbr} method
are conducted on 8 benchmarks.
\textbf{CIFAR10(/100)}~\cite{krizhevsky2009cifar}:
An imagery dataset containing
50,000/10,000 training and testing data
drawn from 10(/100) classes uniformly.
\textbf{SVHN}~\cite{netzer2011svhn}:
The Street View House Numbers dataset
includes 73,257/26,032 train/test images
lying in 10 digit classes $0 \sim 9$.
\textbf{ImageNet}~\cite{russakovsky2015imagenet}:
A large scale imagery dataset
with over 1.2 million images 
from 1,000 classes.
\textbf{CUB}~\cite{WelinderEtal2010cub}:
Caltech-UCSD-Birds contains 11,788 images from 200 breeds of birds
with 312 binary attributes annotations.
\textbf{FLO}~\cite{nilsback2008flo}:
Oxford Flower dataset gathers images from 102 flower categories 
with each class consisting of between 40 and 258 instances.
\textbf{SUN}~\cite{patterson2012sun}:
SUN Attribute is another common fine-grained datasets
used in ZSL with 14,303 images included.
\textbf{AwA2}~\cite{xian2018awa}:
Animals with Attributes2 consists of
37,322 images of 50 animals classes
with 85 numeric attributes for each class.

\paragraph{\bf Experimental setup.}
To transfer knowledge across domains in an unsupervised manner,
we assume that human annotations are only available on source domains
and take the number of target classes as the only prior.
We aim to provide a generic solution
to unsupervised transfer learning 
with fewer assumptions than most of the existing settings.
To that end,
in addition to comparing with transfer clustering techniques
following the same setups as \cite{han2019DTC}
on four benchmarks,
we further evaluated
the effectiveness of \method{abbr} on four ZSL benchmark datasets.
Note, \method{abbr} does not utilise any word-vector embedding space
knowledge on either the source or the target class labels as compared
to the ZSL methods.
We use the same data splits 
as \cite{xian2019f_vaegan_d2}
for fair comparisons.

\paragraph{\bf Performance metrics.}
We adopt two standard metrics in cluster analysis
for evaluation:
(a) Accuracy (\textbf{ACC}) 
is determined by the percentage of test samples 
that are assigned into the cluster
which is matched with the correct ground-truth class with minimum weight.
(b) Normalised Mutual Information (\textbf{NMI})
quantifies the normalised mutual dependence between 
the predicted assignments and the ground-truth memberships.
Both of these metrics are falling within the range of $[0, 1]$
and higher values indicate better performances.
ZSL usually takes Top-1 accuracy as the performance metric,
nevertheless, this is not applicable in our case
due to the deprecation of classes description.
In this case,
we instead report our clustering accuracy.
Although they are not strictly comparable,
they both reveal the model's discrimination ability
and are within the same scale.
Therefore,
we put them in the same tables 
to provide an intuitive comparison.

\paragraph{\bf Implementation details.}
We used the same network architectures
as the ones adopted by \cite{han2019DTC}
as well as the corresponding model weights 
pretrained on source domains provided by them
on the transfer clustering evaluation and
the ImageNet pretrained ResNet101~\cite{chao2016empirical} on ZSL 
to be consistent with~\cite{xian2019f_vaegan_d2}.
The Adam algorithm~\cite{kingma2014adam} 
is adopted for model training
with a fixed learning rate ($1e\!-3\!$).
All the models are randomly initialised and trained with 100 epochs
without $l$2 regularisation.
The image transformations we used for data perturbation
include random rescale and random horizontal flip,
which are also adopted by~\cite{han2019DTC}.
All the main results are averaged over 10 runs
while the ones from ImageNet are averaged over 3 runs with different data splits
following~\cite{han2019DTC,hsu2017KCL,hsu2019MCL}.

\subsection{Unsupervised Transfer Learning}

\paragraph{\bf Transfer Clustering.}
We first evaluated our method's effectiveness 
on clustering unlabelled data
with the help of prior-knowledge from source domains
by comparing with the state-of-the-art transfer clustering models
on four benchmarks.
\begin{table}[h]
\setlength{\tabcolsep}{0.08cm}
\caption{Comparisons on transfer clustering tasks.
The 1st/2nd best results are marked in \best{red}/\second{blue}.
Results of previous methods are adopted from \cite{han2019DTC}.
}
\begin{center}
\begin{tabular}{l|c|c|c|c|c|c|c|c}
\hline
& \multicolumn{2}{c|}{CIFAR-10}
& \multicolumn{2}{c|}{CIFAR-100}
& \multicolumn{2}{c|}{SVHN}
& \multicolumn{2}{c}{ImageNet} \\
\hline
Method & ACC & NMI & ACC & NMI & ACC & NMI & ACC & NMI \\
\hline\hline
K-means 
& 0.655 & 0.422 & 0.662 & 0.555 & 0.426 & 0.182 & 0.719 & 0.713 \\
LPNMF~\cite{cai2009locality}
& - & - & - & - & - & - & 0.430 & 0.526 \\
LSC~\cite{chen2011large}
& - & - & - & - & - & - & 0.733 & 0.733 \\
KCL~\cite{hsu2017KCL}
& 0.665 & 0.438 & 0.274 & 0.151 & 0.214 & 0.001 & 0.738 & 0.750 \\
MCL~\cite{hsu2019MCL}
& 0.642 & 0.398 & 0.327 & 0.202 & 0.386 & 0.138 & \second{0.744} & 0.762 \\
DEC~\cite{xie2016dec}
& 0.749 & 0.572 & 0.721 & 0.630 & 0.576 & 0.348 & \best{0.783} & \second{0.790} \\
DTC~\cite{han2019DTC}
& \second{0.875} & \second{0.735} & \second{0.728} & \second{0.634} & \second{0.609} & \second{0.419} & \best{0.783} & \best{0.791} \\
\hline
\bf SUPREME (ours)
& \best{0.913} & \best{0.791} & \best{0.754} & \best{0.647} & \best{0.787} & \best{0.607} & 0.736 & 0.715 \\
\hline
\end{tabular}
\end{center}
\label{tab:tc_sota}
\end{table}
Results in Table~\ref{tab:tc_sota} show that:
\textbf{(1)}
Most of the unsupervised transfer clustering methods
yield superior performances than K-means.
As a cross-domain deployment solution,
K-means generally applies the knowledge acquired 
from one domain to another
without any selection or adaptation.
Its disadvantages demonstrate 
the necessity of
dealing with distribution shift/discrepancy.
\textbf{(2)}
The proposed \method{abbr} method
surpasses all the competitors
on the first three datasets
mostly with significant margins.
This suggests the effectiveness of 
our auxiliary learning design,
which leverages the self-supervision to 
fill in the gap where
prior-knowledge is not transferrable.
\textbf{(3)}
The advantages of \method{abbr} on ImageNet
is weaker than that on others.
One possible reason is that,
the more complex and diverse image variations 
exhibited in such a larger scale dataset 
lead to more severe intra-class variance and stronger inter-class
similarity. Our self-supervision was constructed 
by relatively limited variations from image transformation,
which may have not provided sufficient inter-sample variations on a
larger scale. 
Nevertheless,
this doesn't degrade the contribution of our key idea
of complementing source domain transferred knowledge with
target domain self-supervision. It encourages to exploit more fully on
larger scale data.

To provide a more intuitive interpretation
of the effects of \method{abbr}, 
we visualise the representations of target samples
in both the pretrained and transferred feature spaces.
\begin{figure*}[ht]
\centering
\includegraphics[width=1.0\linewidth]{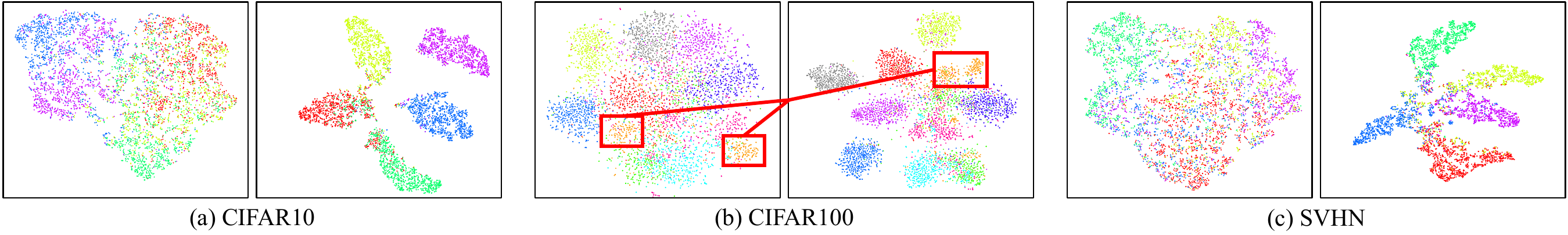}
\caption{
Visualisation of target samples' representation before and after
unsupervised transfer learning by SUPREME.
The left in each pair of images are 
the feature space
produced by the model pretrained on source data,
while the right images are 
by SUPREME.
Ground-truth labels are colour-coded.
}
\label{fig:tsne}
\end{figure*}
Fig~\ref{fig:tsne} shows that,
even though the target samples can roughly 
form some groups in the pretrained feature space,
the internal structure of them are loose
and they are closely adjacent to each other.
As our objective function
encourages determined assignments,
which means no samples should hold similar probabilities
to be assigned into multiple clusters,
our models can yield
clusters in higher compactness and discriminativeness.
Moreover,
according to
the highlighted part in Fig~\ref{fig:tsne}~(b),
thanking to the auxiliary supervision 
constructed by 
intrinsic information on target data,
our method is able to correctly cluster samples of the same classes
but are initially
far away from each other.

\paragraph{\bf Zero-shot Learning.}
In addition to unsupervised image clustering,
we also compared the proposed \method{abbr} method
with ZSL approaches.
The experimental results indicate that our approach to
unsupervised transfer learning provides a plausible generic solution to
other related learning tasks.
\begin{table}[ht!]
\setlength{\tabcolsep}{0.2cm}
\caption{Comparisons on ZSL tasks.
Results of previous works are from~\cite{xian2019f_vaegan_d2}.
}
\begin{center}
\begin{tabular}{l|l|c|c|c|c}
\hline
& Method & CUB & FLO & SUN & AWA2 \\ \hline\hline
\multirow{5}{*}{IND} 
& ALE~\cite{akata2015label}
& 54.9 & 48.5 & 58.1 & 59.9 \\
& CLSWGAN~\cite{xian2018feature}
& 57.3 & 67.2 & 60.8 & 68.2 \\
& SE-GZSL~\cite{kumar2018generalized}
& 59.6 & - & 63.4 & 69.2 \\
& Cycle-CLSWGAN~\cite{felix2018multi}
& 58.6 & 70.3 & 59.9 & 66.8 \\
& F-VAEGAN-D2~\cite{xian2019f_vaegan_d2}
& 61.0 & 67.7 & 64.7 & 71.1 \\
\hline
\multirow{6}{*}{TRAN} 
& ALE-tran~\cite{xian2018awa}
& 54.5 & 48.3 & 55.7 & 70.7 \\
& GFZSL~\cite{verma2017simple}
& 50.0 & 85.4 & 64.0 & 78.6 \\
& DSRL~\cite{ye2017zero}
& 48.7 & 57.7 & 56.8 & 72.8 \\
& UE-finetune~\cite{song2018qfsl}
& 72.1 & - & 58.3 & 79.7 \\
& F-VAEGAN-D2~\cite{xian2019f_vaegan_d2}
& 71.1 & 89.1 & 70.1 & 89.8 \\ \hline
& \textbf{SUPREME (ours)}
& 68.3 & 67.1 & 41.3 & 76.3 \\
\hline
\end{tabular}
\end{center}
\label{tab:tzsl_sota}
\end{table}
Table~\ref{tab:tzsl_sota} shows that
transductive learning methods for ZSL are likely to produce higher accuracy
by leveraging additional unlabelled target data
in model learning,
which is consistent with our training strategy.
The competitive results yielded by
\method{abbr}
on all the datasets 
except SUN
demonstrate its discrimination ability
on target domains
even without any word vector prototype mapping in the text space or
human labelled attribute learning on class description.
However,
the rather poor performance on SUN,
in which the number of target samples is limited to around 1,000
but the target classes are larger than other benchmarks,
implies that sufficient training data  
is still required by our model 
even without labels.

\subsection{Component Analysis and Discussions}

We conducted detailed ablation studies to investigate
the effectiveness of different design choices in our model
for in-depth analysis.

\paragraph{\bf Transferred supervision v.s. Self-supervision.}
As our key idea holds the assumption that
self-supervision can provide
complementary constraints to 
the target samples 
to which the prior-knowledge is not applicable,
we evaluated the effectiveness of 
both the transferred supervision and self-supervision
to better understand their individual contributions to the model.
%
\begin{figure}[ht]
\centering
\includegraphics[width=0.95\linewidth]{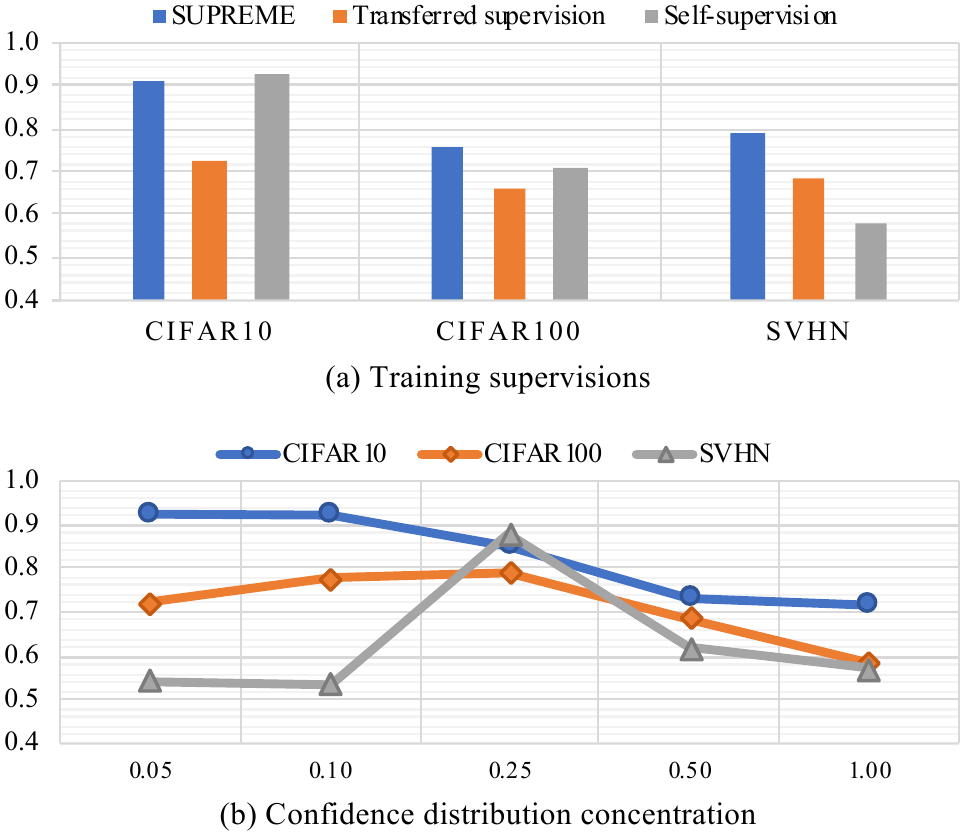}
\caption{
Ablation studies of training objective.
\textbf{(a)} Decouple different supervisions 
to investigate their individual contribution to model's capacity.
\textbf{(b)} Effect of temperature $\tau$ 
which decides the concentration degree of confidence distribution.
}
\label{fig:supervision_confidence}
\end{figure}
According to Fig~\ref{fig:supervision_confidence} (a),
both the prior-knowledge and the self-supervision
are able to provide useful constraints independently in model training.
Our \method{abbr} model achieves better performance in most cases,
which indicates 
these two supervisions can benefit each other.
It is also interesting to see that the self-supervised model marginally surpasses
\method{abbr} on CIFAR10.
This suggests that the prior-knowledge from the source domain sometimes contains
misleading and inaccurate (nontransferrable) information of high confidence to the target domain. 

\paragraph{\bf Confidence Distribution Concentration.}
The temperature $\tau$ used to
compute the confidence of constraints 
in Eq~\ref{eq:certainty}
decides the concentration degree of 
the normalised confidence distribution,
hence,
it can be interpreted as the 
reliability of prior-knowledge 
in terms of self-supervision.
As shown in Fig~\ref{fig:supervision_confidence} (b),
our model is able to attain promising performance 
with a wide range of $\tau$
but the best result is sometimes achieved 
at different values on different datasets.
This is actually within expectation because
measuring the transferability of prior-knowledge
is intrinsically challenging
and the setting of $\tau$ is intricately related to various factors,
\textit{e.g.} capacity of pretrained models.
Due to the existence of human annotations on source domains,
a reasonable setting of $\tau$ can be determined by cross-validation,
which is a conventional solution in transfer learning~\cite{xian2018awa}.

\paragraph{\bf Regularisations.}
\begin{figure}[ht]
\centering
\includegraphics[width=0.95\linewidth]{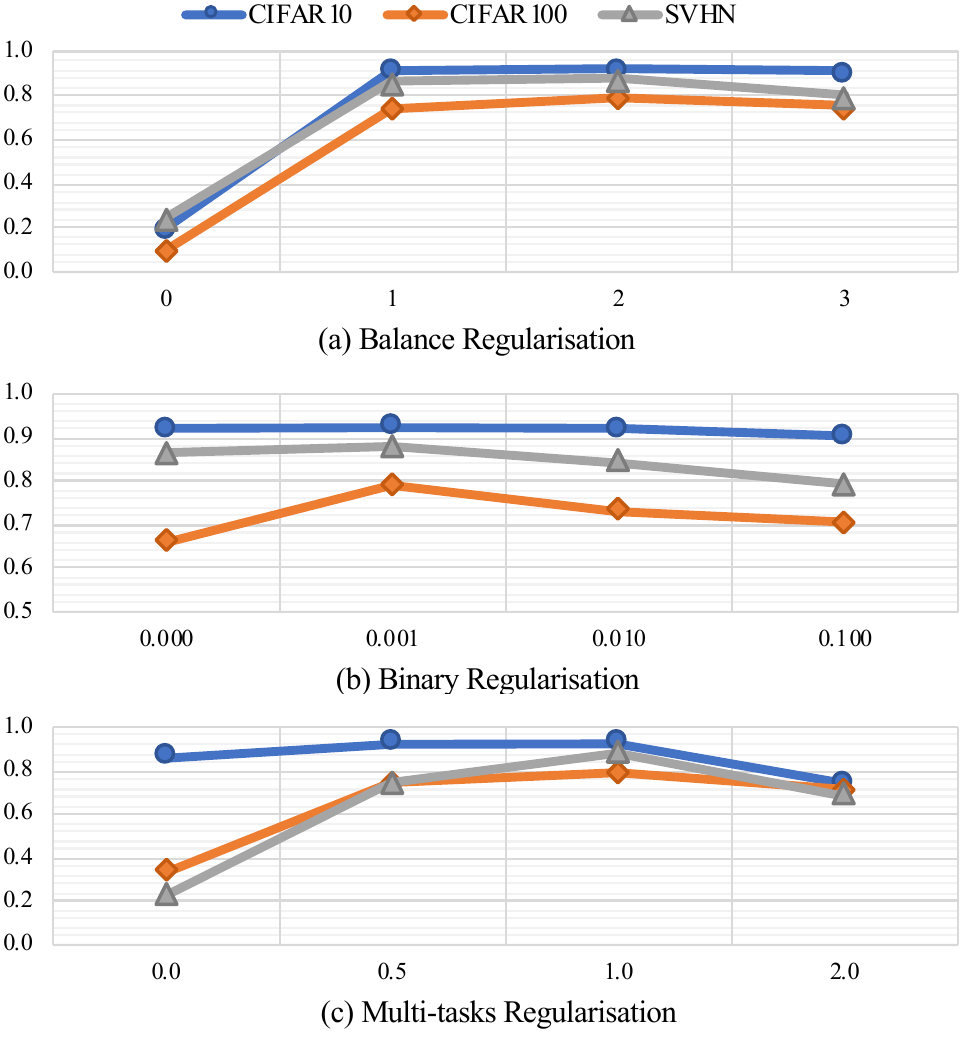}
\caption{
Effects of different regularisations. 
\textbf{(a)} Balance Regularisation
on clusters size distribution.
\textbf{(b)} Binary Regularisation
on common factor space.
\textbf{(c)} Multi-tasks Regularisation
to avoid trivial representation.
}
\label{fig:regularisations}
\end{figure}
%
The \method{abbr} model is trained with
several regularisations
and
we investigated their necessity
as well as 
our model's robustness to them
by varying their weights within different ranges.
As shown in Fig~\ref{fig:regularisations},
the significant performance drop in most cases 
caused by removing either of these regularisations
(setting the weight to $0$)
demonstrate that 
none of them is redundant for effective knowledge transfer.
Furthermore,
the stability and the similar trends in parameter values on different datasets
indicate the 
scalability and 
robustness of our model
which requires no exhaustive parameter tuning.

\section{Conclusion}\label{sec:conclude}

In this work,
we addressed a common underlying problem among several unsupervised
transfer learning approaches that aim to model a discriminative latent space 
in an unsupervised manner 
with the help of the prior-knowledge acquired from related domains.
To that end, we propose a more scalable solution to unsupervised
transfer learning by formulating a \textit{\method{full}} (\method{abbr}) method
to augment
transfer clustering with self-supervised learning.
It is inevitable that
some of the target samples will fail to yield reliable transferred supervision from 
prior-knowledge due to distribution shift/discrepancy,
the proposed \method{abbr} method is designed to identify those target
samples and provide them with self-supervision based on intrinsic
pairwise similarities among the target images in relation to the
source domain pre-knowledge.
Extensive experiments on four transfer clustering benchmarks
demonstrate the superiority of the proposed method
over a wide range of the state-of-the-art models.
Moreover, \method{abbr} also shows competitive discrimination ability
on ZSL benchmarks 
without utilising any additional semantic knowledge
representation either in the word to vector text space or attribute
learning. 
Ablation studies
and in-depth analysis 
are conducted
to give insights on \method{abbr} design considerations.






\bibliographystyle{IEEEtran}
\bibliography{press_full,egbib}
\end{document}